\documentclass[11pt,a4paper]{article}
\usepackage[T1]{fontenc}
\usepackage[utf8]{inputenc}
\usepackage{acl}
\usepackage{times}
\usepackage{latexsym}

\usepackage{graphicx}
\usepackage{booktabs}
\usepackage{microtype}
\usepackage{amsmath}
\usepackage{amssymb}
\usepackage{multirow}
\usepackage{tabularx}
\usepackage{enumitem}
\setlist{nosep,topsep=0pt}
\usepackage{fancyvrb}
\usepackage[table]{xcolor}
\usepackage{colortbl}
\usepackage[disable]{todonotes} 
\usetikzlibrary{arrows.meta,positioning,shapes.geometric}
\hypersetup{pdfauthor={}, pdftitle={}, pdfsubject={}, pdfkeywords={}}
\definecolor{CDDPrimaryBlue}{HTML}{2E5C8A}
\definecolor{CDDAccentOrange}{HTML}{C66B3D}
\definecolor{CDDNeutralGray}{HTML}{6B6B6B}
\definecolor{CDDLightGray}{HTML}{D9D9D9}
\definecolor{CDDErrorRed}{HTML}{B33A3A}
\definecolor{primaryblue}{HTML}{2E5C8A}
\definecolor{accentorange}{HTML}{C66B3D}
\definecolor{neutralgray}{HTML}{6B6B6B}
\definecolor{lightgray}{HTML}{D9D9D9}
\definecolor{errorred}{HTML}{B33A3A}
\definecolor{baseline1}{HTML}{BDBDBD}
\definecolor{baseline2}{HTML}{969696}
\definecolor{baseline3}{HTML}{7F7F7F}
\definecolor{baseline4}{HTML}{666666}
\definecolor{baseline5}{HTML}{4D4D4D}

\title{Does RAG Know When Retrieval Is Wrong?\\ Diagnosing Context Compliance under Knowledge Conflict}

\author{
Yihang Chen\textsuperscript{1,*}\quad
Pin Qian\textsuperscript{2,*}\quad
Su Wang\textsuperscript{2,*}\quad
Sipeng Zhang\textsuperscript{3,*}\\
Huan Xu\textsuperscript{1}\quad
Shuhuai Lin\textsuperscript{2}\quad
Xinpeng Wei\textsuperscript{1,\textdagger}\\[0.25em]
{\normalfont\small
\textsuperscript{1}Georgia Institute of Technology\quad
\textsuperscript{2}Carnegie Mellon University\quad
\textsuperscript{3}University of California San Diego}\\[0.2em]
{\normalfont\footnotesize\texttt{ychen3726@gatech.edu, pqian@alumni.cmu.edu}}\\
{\normalfont\footnotesize\texttt{suwang@alumni.cmu.edu, siz018@ucsd.edu, huan.xu71@gmail.com}}\\
{\normalfont\footnotesize\texttt{shuhuail@andrew.cmu.edu, william.xp.wei@outlook.com}}\\[0.2em]
{\normalfont\footnotesize
\textsuperscript{*}Equal contribution.\quad
\textsuperscript{\textdagger}Corresponding author.}
}

\date{}

\begin{document}
\maketitle

\begin{abstract}
Retrieval-Augmented Generation (RAG) is usually evaluated by whether the final answer is correct. Under knowledge conflict, this hides a key question: did the model follow retrieved evidence, rely on its parametric prior, or produce a post-hoc rationale? We study this as context compliance, the regime in which retrieved context controls the answer even when it conflicts with the model's prior knowledge. We introduce Context-Driven Decomposition (CDD), an inference-time diagnostic intervention that elicits contextual and prior answers, isolates the conflicting premise, and records a resolution trace that can be perturbed. Across Epi-Scale stress tests, TruthfulQA misconception injection, and cross-model reruns, CDD makes three behaviors visible. First, misleading retrieval can severely degrade accuracy: under worst-case injected context (a TruthfulQA misconception-injection probe), Standard RAG reaches only 15.0\%. Second, better answers need not share the same mechanism: CDD improves adversarial accuracy on Gemini-2.5-Flash and shows directional gains across Claude variants, yet trace-perturbation sensitivity is high only on Gemini. Third, explicit decomposition improves controlled-conflict robustness over a conflict-aware instruction baseline on localized factual conflicts, with the clearest margins on Entity Swap (88.0\% vs 79.3\%) and Logical Contradiction (83.2\% vs 75.4\%). We frame RAG conflict handling as an observability problem.
\end{abstract}

\section{Introduction}
Standard Retrieval-Augmented Generation (RAG) \citep{lewis2020retrieval} is built on a useful default assumption: retrieved evidence should help the model answer. The assumption becomes fragile when retrieval returns evidence that is wrong or adversarially injected \citep{xie2023adaptive, zou2024poisonedrag}. Consider a query about whether cracking knuckles causes arthritis. If the retrieved context asserts this common misconception \citep{lin2022truthfulqa}, a RAG model that answers ``yes'' may be faithfully following the context while becoming factually wrong; a model that answers ``no'' may have detected the conflict, ignored retrieval, or simply relied on memorized prior knowledge. The final answer alone cannot tell these cases apart.

This is the central blind spot we study. Under knowledge conflict, standard accuracy evaluation collapses several distinct behaviors into the same score, paralleling broader concerns that accuracy-only evaluation can hide trace-level differences \citep{turpin2023,lanham2023measuring,sun2026beyond}. A correct answer does not reveal whether the model used the right source of evidence, rejected a false retrieved premise, or produced a post-hoc rationale. An incorrect answer does not distinguish ordinary knowledge failure from \textbf{context compliance}: retrieved context controlling the answer despite conflicting with the model's parametric prior. RAG therefore needs diagnostics that expose not only \emph{what} the model answered, but \emph{which source controlled the answer} and whether the final answer is sensitive to corruption of that trace.

We introduce \textbf{Context-Driven Decomposition (CDD)} as a diagnostic intervention for this setting. CDD decomposes a conflicted RAG instance into five observable steps: elicit the answer implied by the retrieved context, elicit the model's parametric answer, compare them, isolate the conflicting premise, and resolve the conflict. This turns an implicit arbitration problem into a trace that can be tested for causal influence---and we find that influence is model-specific. We then perturb that trace with mistake-injection and truncation tests to ask whether the trace actually affects the final answer. CDD is not presented as a universal production defense; its role is to make hidden conflict-resolution behavior observable and testable.

We make two methodological contributions and one empirical contribution:
\begin{enumerate}[label=\arabic*.]
    \item \textbf{Conflict observability:} We operationalize context compliance at the prompt level and show why final-answer accuracy cannot identify whether retrieved context, parametric knowledge, or a post-hoc rationale controlled the output.
    \item \textbf{Diagnostic intervention:} We introduce CDD, a decomposition-based probe that exposes contextual belief, parametric belief, conflict detection, premise isolation, and resolution, then tests whether perturbing the resolution trace changes the answer.
    \item \textbf{Empirical finding:} We show that CDD improves controlled-conflict robustness while revealing that accuracy transfer and trace-perturbation sensitivity can dissociate across model families. We will release the Epi-Scale split (2,250 clean + 2,250 adversarial records), prompt templates, evaluation settings, and scoring specifications upon publication; the internal experimental codebase is not released.
\end{enumerate}

\textbf{Scope.} The scope of this paper is diagnostic. We use CDD to expose when conflict-resolution behavior is present, absent, or model-specific. The present evidence should be read as a controlled study rather than a deployment recommendation: organic multi-document retrieval, stronger faithfulness interventions, and open-weight replications remain follow-up evaluations.

\section{Related Work}

\textbf{Knowledge conflict in RAG.} The tension between parametric memory and external evidence is well documented. Prior work studies when models should rely on parametric versus retrieved knowledge \citep{mallen2023}, how entity substitutions alter QA outputs \citep{longpre2021}, and how models respond to conflicting retrieved evidence \citep{xie2023adaptive,chen2022rich}. These studies establish that conflict changes model behavior; our focus is the next diagnostic question: when a final answer changes under conflict, can we observe which source controlled it and whether the final answer is sensitive to corruption of that trace?

Recent work has refined this picture along three axes. ClashEval \citep{wu2024clasheval} systematically quantifies the ``tug-of-war'' between an LLM's parametric prior and external evidence as a function of evidence quality, framing the conflict as a measurable spectrum. Astute RAG \citep{wang2024astute} addresses imperfect retrieval and knowledge conflict at the method level by combining internal-knowledge elicitation with iterative consolidation. Corrective Retrieval-Augmented Generation (CRAG) \citep{yan2024corrective} inserts a retrieval evaluator that classifies retrieved evidence into correct/ambiguous/incorrect bins before generation, modifying the retrieval pipeline rather than the generation step. A recent survey of knowledge conflicts in LLMs \citep{xu2024knowledgeconflicts} organizes failure modes into context-memory, inter-context, and intra-memory conflicts. CDD targets the context-memory conflict axis specifically and treats conflict not only as a failure to mitigate, but as a situation in which the model's source arbitration should be made observable.

\textbf{Robust and filtered RAG.} Self-RAG \citep{asai2024selfrag} trains models to generate reflection tokens, while prior document-filtering RAG removes unsupported retrieved documents before generation \citep{yoran2023making}. In our experiments, the reported reflection baseline is a prompt-only approximation inspired by Self-RAG rather than the trained Self-RAG model itself. These methods attempt to improve generation quality or retrieval reliability. CDD differs in objective: it keeps the conflict visible long enough to ask whether the model detected, rejected, or complied with the retrieved premise. Our filtering baseline is likewise prompt-only: it instructs the model to remove implausible claims from the context, and does not use a trained NLI model.

\textbf{Retrieval routing, filtering, and deployment trade-offs.} CDD-$\alpha$ is a diagnostic routing rule rather than a production retrieval architecture, but related routing and filtering ideas appear in broader RAG and inference systems. Domain-specific RAG work uses reranking or document routing to manage evidence selection and robustness--precision trade-offs in financial-document QA \citep{cheng2026enhancingfinancialreportquestionanswering,cheng2026resolvingrobustnessprecisiontradeofffinancial}. Outside the specific knowledge-conflict setting studied here, sequential filtering provides a methodologically adjacent model of how staged filters trade off cost and selectivity \citep{paranjape2026optimality}. On-device RAG benchmarks are a complementary deployment consideration, showing that retrieval, reranking, and generation also create latency and energy constraints \citep{cheng2026energyefficientondeviceragmobile}. These works motivate compute-aware routing as an operating point; CDD uses routing only to decide when to invoke a conflict diagnostic trace.

\textbf{Context-aware decoding and parametric-contextual conflict.} Context-Aware Decoding contrasts contextual and parametric logits to reduce hallucination and increase reliance on retrieved evidence \citep{shi2023trusting}. Our setting highlights a complementary concern: if retrieved evidence is itself false, more context reliance can be the wrong behavior. We target closed API models where logit-level access is unavailable; instead of computing token-level divergence directly, CDD elicits contextual and parametric answers in natural language and treats the resulting trace as an observable diagnostic artifact.

\textbf{Faithfulness of chain-of-thought.} A critical challenge in chain-of-thought (CoT) reasoning is whether the generated rationale actually influences the final answer. Prior work shows that CoT explanations can be systematically misleading \citep{turpin2023}. Causal-intervention work introduces tests including early answering and adding mistakes \citep{lanham2023measuring}. We adapt these ideas to retrieval-induced conflict; in this paper, \textit{Truncation} refers to the early-answering-style intervention on the CDD trace, and \textit{Mistake Injection} refers to the adding-mistakes-style intervention. This connection is central: a decomposition trace is useful as a diagnostic only if we can test whether perturbing it changes the answer.

\textbf{Evaluation reliability and trace diagnostics.} Our causal-sensitivity tests are also motivated by broader evaluation context: conclusions can depend on what artifact is scored and how the benchmark is configured. LLM-as-a-judge surveys document the scope and reliability challenges of judge-based evaluation \citep{li2025generation}; trace-level evaluation work shows that identical final-answer scores can mask different intermediate behavior \citep{sun2026beyond}. Adjacent trace-audit studies further show that parser, prompt-template, or probe choices can affect conclusions drawn from reasoning or memorization traces \citep{li2026auditingreasoningtracememorizationclaims,fan2026probe}. Other methodologically adjacent evaluation work makes the same caution concrete in different settings: regime-stratified benchmarking shows that aggregate metrics can hide operating-regime failures \citep{wang2026timeseriesfoundationmodel}, alignment-benchmark audits show that ranks can be configuration-conditional \citep{li2026safetyreproconfigurationconditionalrankinstability}, and paired-MDE planning argues for specifying detectable effects before interpreting small differences \citep{zhuang2026preregisteringdetectableeffectpairedmde}. We therefore treat CDD traces as diagnostic artifacts whose causal role must be tested, not as explanations to be trusted at face value.

\begin{figure*}[t]
\centering
\includegraphics[width=0.98\textwidth]{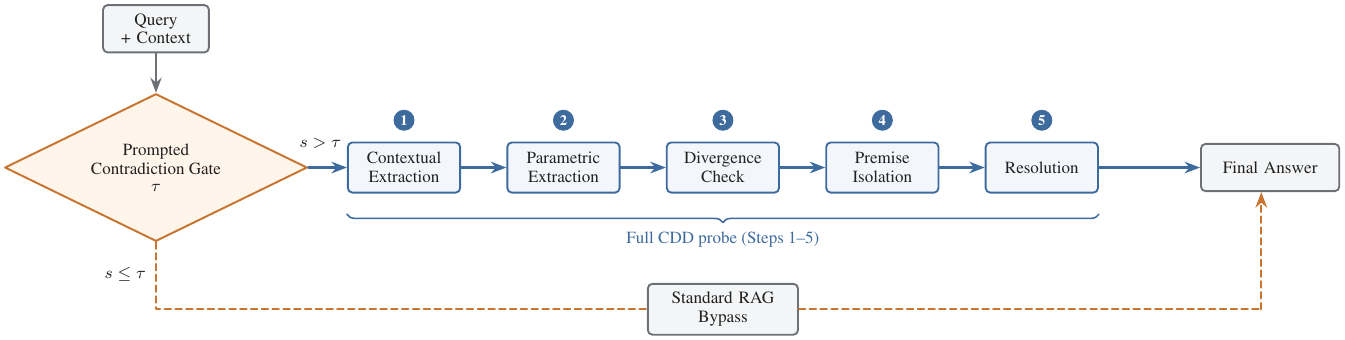}
\caption{CDD pipeline with the CDD-$\alpha$ prompted contradiction gate. High-conflict samples enter the full decomposition probe, while low-conflict samples follow the Standard RAG bypass before converging at the final answer.}
\label{fig:pipeline}
\end{figure*}

\section{A Belief-Revision Framework for Context-Parametric Conflict}

\subsection{Conceptual Formulation of Conflict}
We borrow the vocabulary of belief revision \citep{agm1985} as a conceptual scaffold; the 1985 formalism concerns logical theory change and makes no claim about LLM internals. We distinguish between the model's parametric prior $P_\theta(a \mid q)$---internal world knowledge---and its contextual posterior $P_\theta(a \mid q, c)$ conditioned on retrieved evidence. The \textbf{Compliance Regime} arises when the posterior dominates despite divergence from the prior; standard RAG can enter this regime because generation is directly conditioned on retrieved evidence. The \textbf{Resolution Regime}, in contrast, requires detection and arbitration of divergence. While the conceptual axis between these regimes corresponds to a divergence measure between the two distributions (e.g., Jensen-Shannon Divergence), exact token-level computation requires white-box logit access. Our prompt-based instantiation makes this divergence observable through explicit answer generation rather than logit comparison; we treat token-level JSD instantiation on open-weight models as a natural extension. In our protocol, Step 2 elicits the parametric answer with the context already visible; it is therefore a prior reported under the CDD prompt, not a context-free parametric answer. The belief-revision formulation is a conceptual scaffold, not an instantiated measurement.

\subsection{The CDD Framework}
CDD makes the implicit conflict explicit via a five-step reasoning trace (Figure~\ref{fig:pipeline}). These five steps function as a probe instrument: each step exposes a specific belief-revision operation that remains implicit under standard RAG.
\begin{enumerate}[label=\textbf{Step \arabic*:}]
    \item \textbf{Contextual Extraction:} Output $\hat{a}_{ctx}$.
    \item \textbf{Parametric Extraction:} Output $\hat{a}_{param}$.
    \item \textbf{Divergence Check:} Compare $\hat{a}_{ctx}$ and $\hat{a}_{param}$.
    \item \textbf{Premise Isolation:} If they conflict, isolate the premises causing the conflict and evaluate whether they violate established history or science.
    \item \textbf{Resolution:} Resolve the conflict and output the final answer.
\end{enumerate}

\subsection{Algorithmic Variant: CDD-\texorpdfstring{$\alpha$}{alpha}}
We also report \textbf{CDD-$\alpha$}, a compute-aware routing variant used to study compute/accuracy trade-offs. The context $c$ is paired with the current parametric answer estimate and scored by a prompted Gemini-2.5-Flash contradiction gate, which asks whether the premise explicitly contradicts the hypothesized answer and returns a scalar score in $[0,1]$. If the score exceeds $\tau$, the sample is routed to the deep CDD logic. Otherwise, it defaults to Standard RAG. This routing rule operationalizes a selective-intervention setting in which only high-conflict examples invoke the full decomposition trace; it is not a trained NLI checkpoint or a calibrated probability model.

\section{Experimental Setup}

\subsection{The Epi-Scale Benchmark}
Epi-Scale contains 2,250 base examples drawn evenly from \textbf{HotpotQA} (multi-hop) \citep{yang2018hotpotqa}, \textbf{Natural Questions} (single-hop) \citep{kwiatkowski2019natural}, and \textbf{FEVER} (fact verification) \citep{thorne2018fever}. Each base example is instantiated in two forms: a clean-context record and an adversarial-context record, yielding 4,500 evaluation records in total.

\textbf{Construction:} The clean records use the original retrieved context. The adversarial records are generated through an LLM-based perturbation engine parameterized for high semantic variance. We generated four mutually exclusive perturbation subsets: Entity Swap, Temporal Shift, Logical Contradiction, and Distractor Evidence. (Generation procedure summarized in Appendix~\ref{sec:appendix_data}).

Because the primary evaluation model also generated the perturbations and serves as the CDD-$\alpha$ contradiction gate, we cannot rule out a same-model bias; the Claude-family reruns, which evaluate on perturbations generated by a different model, partly address this.

\textbf{Limitations of Synthetic Perturbations:} While Epi-Scale improves upon templated datasets, LLM-generated adversarial texts often exhibit lower perplexity and uniform lexical diversity compared to organic human misinformation. We mitigate this by including a real-world evaluation on \textbf{TruthfulQA} \citep{lin2022truthfulqa}.

\subsection{Evaluation Scope}
We use three evaluation settings, each with a different diagnostic role. Epi-Scale synthetic conflict is a controlled perturbation stress test for isolating specific conflict types. TruthfulQA misconception injection is a worst-case upper-bound compliance test, not an organic retrieval benchmark. Claude-family replication is a cross-model diagnostic check for whether the observed coupling and conflict-resolution signals are model-specific. The main adversarial analysis reports results on the Epi-Scale adversarial records: 2,250 examples evenly divided across the four perturbation types ($\sim$562--563 per type). Per-perturbation accuracies in Table~\ref{tab:adversarial} are computed over all examples in each group, and the macro average is the unweighted arithmetic mean across the four perturbation cells.

\subsection{Evaluation Metrics}
We use normalized string matching (lowercase, punctuation stripping, alias mapping) to reduce superficial form mismatches. Hedged or non-committal responses are handled by task-specific rules described in Appendix A.
\begin{itemize}
    \item \textbf{Accuracy and macro average:} Per-perturbation accuracy is computed independently for each perturbation group. ``Macro Avg.'' denotes the unweighted arithmetic mean of the four displayed perturbation accuracies.
    \item \textbf{Confidence intervals:} For Table~\ref{tab:adversarial}, each per-cell 95\% CI is a normal-approximation binomial interval of half-width $z_{0.975}\sqrt{\hat{p}(1-\hat{p})/n}$ over the $\sim$562--563 examples in that perturbation group. The macro-average half-width is computed as $z_{0.975}\times(1/4)\times\sqrt{\sum_i \hat{p}_i(1-\hat{p}_i)/n_i}$, under the assumption that the four perturbation cells are independent samples. We note that this propagation does not capture potential correlations introduced by the shared question pool from HotpotQA/NQ/FEVER.
    \item \textbf{Causal Sensitivity:} We quantify faithfulness using intervention tests (Truncation and Mistake Injection) \citep{lanham2023measuring}. Sensitivity is the relative accuracy drop, defined as $(\text{Acc}_{\text{clean}}-\text{Acc}_{\text{corrupted}})/\text{Acc}_{\text{clean}}$.
\end{itemize}

\subsection{Significance Reporting}
In place of paired hypothesis tests, we report a CI-based conservative significance check on the largest method gaps. For the Epi-Scale adversarial records, we report per-cell binomial 95\% confidence intervals. The CIs for CDD and a strong non-CDD baseline (Reflection Prompting) do not overlap on Entity Swap (88.0\% $\pm$2.7 vs 69.5\% $\pm$3.8) or Logical Contradiction (83.2\% $\pm$3.1 vs 65.0\% $\pm$3.9). Conflict-aware RAG is a stronger instruction-matched baseline (73.5 macro) over which CDD retains a margin on Entity Swap and Logical Contradiction. The conflict-aware per-cell CIs (Table~\ref{tab:adversarial}) do not overlap with CDD on Entity Swap or Logical Contradiction, while the Temporal Shift and Distractor Evidence intervals overlap. Mixed clean/adversarial comparisons use paired bootstrap over the paired 4,500-record Epi-Scale evaluation set, with the harmonic mean of clean and adversarial accuracy as the aggregate robustness metric.

\subsection{Model Setup}
We use Gemini-2.5-Flash for primary evaluations. To study cross-architecture generalization, we extend evaluations to Claude Haiku, Sonnet, and Opus; exact API model identifiers are listed in the reproducibility statement. Closed-API behavior may still drift on non-pinned dependencies, which we treat as a reproducibility limitation.

\section{Results}

\begin{table*}[t]
\centering
\caption{Adversarial accuracy across perturbation types on the Epi-Scale adversarial records (Gemini-2.5-Flash, N=2,250; $\sim$562--563 samples per perturbation; normal-approximation 95\% binomial CIs). Macro Avg. is the unweighted arithmetic mean of the four perturbation accuracies; the rightmost interval propagates the four per-cell binomial uncertainties.}
\label{tab:adversarial}
\small
\setlength{\tabcolsep}{3.2pt}
\renewcommand{\arraystretch}{0.95}
\rowcolors{2}{gray!6}{white}
\begin{tabular}{@{}lccccc@{}}
\toprule
\textbf{Model \& Setting} & \textbf{Entity Swap} & \textbf{Log. Contradict.} & \textbf{Temp. Shift} & \textbf{Distract. Evid.} & \textbf{Macro Avg.} \\
\midrule
ClosedBook (Zero-shot) & 43.7\% \scriptsize($\pm$4.1) & 40.7\% \scriptsize($\pm$4.1) & 50.0\% \scriptsize($\pm$4.1) & 44.4\% \scriptsize($\pm$4.1) & 44.7\% \scriptsize($\pm$2.0) \\
Standard RAG & 58.4\% \scriptsize($\pm$4.1) & 56.0\% \scriptsize($\pm$4.1) & 68.8\% \scriptsize($\pm$3.8) & 68.8\% \scriptsize($\pm$3.8) & 63.0\% \scriptsize($\pm$2.0) \\
Vanilla CoT & 62.0\% \scriptsize($\pm$4.0) & 61.3\% \scriptsize($\pm$4.0) & 63.2\% \scriptsize($\pm$4.0) & 68.1\% \scriptsize($\pm$3.9) & 63.7\% \scriptsize($\pm$2.0) \\
Reflection Prompting & 69.5\% \scriptsize($\pm$3.8) & 65.0\% \scriptsize($\pm$3.9) & 66.0\% \scriptsize($\pm$3.9) & 67.5\% \scriptsize($\pm$3.9) & 67.0\% \scriptsize($\pm$1.9) \\
Prompt-Filtered RAG & 68.0\% \scriptsize($\pm$3.9) & 64.5\% \scriptsize($\pm$4.0) & 65.5\% \scriptsize($\pm$3.9) & 67.0\% \scriptsize($\pm$3.9) & 66.3\% \scriptsize($\pm$2.0) \\
Conflict-aware RAG & 79.3\% \scriptsize($\pm$3.3) & 75.4\% \scriptsize($\pm$3.6) & 70.0\% \scriptsize($\pm$3.8) & 69.2\% \scriptsize($\pm$3.8) & 73.5\% \scriptsize($\pm$1.8) \\
\rowcolor{primaryblue!15} 
\textbf{CDD (Ours)} & \textbf{88.0\%} \scriptsize($\pm$2.7) & \textbf{83.2\%} \scriptsize($\pm$3.1) & \textbf{71.3\%} \scriptsize($\pm$3.7) & \textbf{69.9\%} \scriptsize($\pm$3.8) & \textbf{78.1\%} \scriptsize($\pm$1.7) \\
\bottomrule
\end{tabular}
\rowcolors{2}{}{}
\end{table*}

\begin{table}[t]
\centering
\caption{Cross-model replication on the Epi-Scale adversarial records (Claude-family models, N=2,250 adversarial examples). CDD improves adversarial accuracy on all three Claude-family models. The per-model gains are directional point estimates (+2.2/+4.6/+2.6 pp). CDD marginal 95\% confidence intervals are shown in the table; paired significance tests for the method differences were not run. This reflects transfer of the accuracy benefit only; trace-perturbation sensitivity transfer is reported as point estimates without paired significance tests (\S5.5.3).}
\label{tab:cross_model}
\scriptsize
\setlength{\tabcolsep}{2.8pt}
\renewcommand{\arraystretch}{0.95}
\rowcolors{2}{gray!6}{white}
\begin{tabular}{@{}lccc@{}}
\toprule
\textbf{Method} & \textbf{Haiku} & \textbf{Sonnet} & \textbf{Opus} \\
\midrule
Standard RAG & \textbf{79.0\%} & \textbf{76.0\%} & 79.4\% \\
Vanilla CoT & 73.6\% & 73.2\% & 75.4\% \\
\rowcolor{primaryblue!15}
\textbf{CDD (Ours)} & \textbf{81.2\%} \tiny($\pm$1.6) & \textbf{80.6\%} \tiny($\pm$1.6) & \textbf{82.0\%} \tiny($\pm$1.6) \\
\bottomrule
\end{tabular}
\rowcolors{2}{}{}
\end{table}

\subsection{Adversarial Stress Test and Baselines}
We compare CDD against existing baselines to establish that decomposition changes behavior under controlled conflict. The comparatively high Claude Standard RAG adversarial accuracies are consistent with perturbations generated by Gemini-2.5-Flash transferring imperfectly across model families, reinforcing the same-model-bias caveat in \S4.1. This is a prerequisite for the later diagnostic question: once behavior changes, does the explicit conflict-resolution trace actually control the answer?

Figure~\ref{fig:adversarial_accuracy} (in Appendix B) visualizes the perturbation-level pattern, while Table~\ref{tab:adversarial} reports the exact values for CDD, chain-of-thought prompting (CoT) \citep{wei2022cot}, reflection-token prompting inspired by Self-RAG \citep{asai2024selfrag}, and a prompt-only self-filtering baseline inspired by document-filtering RAG \citep{yoran2023making} on the Epi-Scale adversarial records (N=2,250) using Gemini-2.5-Flash. The reflection baseline is prompt-only and should not be read as a reproduction of the trained Self-RAG system.

Standard RAG reaches a 63.0\% macro average under targeted misinformation across the adversarial records. CDD reaches 78.1\% macro, with the largest gains on the localized factual conflicts (Entity Swap, Logical Contradiction; Table~\ref{tab:adversarial}); the Temporal Shift and Distractor Evidence intervals overlap with Standard RAG, so we do not claim significance there. A conflict-aware prompt baseline---instructing the model to use the context but override it when it conflicts with established facts---reaches 73.5\% macro, confirming that an explicit conflict-checking instruction accounts for much of the gain; CDD's remaining margin concentrates on Entity Swap (88.0 vs 79.3) and Logical Contradiction (83.2 vs 75.4), the localized factual-conflict settings where premise isolation (Step 4) applies, and collapses on Temporal Shift and Distractor Evidence (overlapping). This pattern indicates that explicit conflict decomposition improves robustness on the localized factual-conflict families (Entity Swap, Logical Contradiction) on Gemini-2.5-Flash, while showing no separable gain on Temporal Shift or Distractor Evidence.

\textbf{Average-case calibration.} In a separate mixed clean/adversarial calibration run over the paired clean and adversarial records, CDD and Standard RAG are statistically tied (72.23\% and 72.33\% harmonic means, respectively; bootstrap p=0.53). This tie hides a trade-off in the underlying components: CDD scores 67.2\% clean / 78.1\% adversarial, whereas Standard RAG scores 84.9\% clean / 63.0\% adversarial. CDD therefore trades roughly 18 points of clean-sample accuracy for its adversarial robustness gain, confirming that CDD is not a universally better answer prompt but a conflict-specific intervention whose value appears only when the retrieved context and the model prior disagree. The higher CDD adversarial score likely reflects that many perturbations alter a target premise while leaving other answer evidence recoverable; normalized-string scoring may also credit partial or hedged answers. We therefore treat Epi-Scale as a controlled stress test, not as a claim that adversarial contexts remove all answer-relevant evidence.

\begin{table}[t]
\centering
\caption{Component-wise ablation on Gemini-2.5-Flash. Entity Sw. and Log. Con. are shown because they are the localized factual-conflict families where CDD's strongest baseline margins occur in Table~\ref{tab:adversarial}. Overall is the unweighted macro average over all four perturbation types (Entity Swap, Logical Contradiction, Temporal Shift, Distractor Evidence). The omitted Temporal Shift and Distractor Evidence cells are not individually recoverable from the archived aggregate table; their two-cell mean is implied by the Overall column.}
\label{tab:ablation}
\scriptsize
\setlength{\tabcolsep}{2.8pt}
\renewcommand{\arraystretch}{0.95}
\rowcolors{2}{gray!6}{white}
\begin{tabular}{@{}lccc@{}}
\toprule
\textbf{Ablation Variant} & \textbf{Entity Sw.} & \textbf{Log. Contradict.} & \textbf{Overall} \\
\midrule
\rowcolor{primaryblue!15}
\textbf{Full CDD} & \textbf{88.0\%} & \textbf{83.2\%} & \textbf{78.1\%} \\
Length-Matched Sham CoT & 42.4\% & 32.0\% & 40.1\% \\
w/o Step 4 (Isolation) & 75.2\% & 72.0\% & 65.1\% \\
w/o Step 3 (Diverge) & 77.0\% & 73.5\% & 66.0\% \\
\bottomrule
\end{tabular}
\rowcolors{2}{}{}
\end{table}

\subsection{Ablation Study}
Table~\ref{tab:ablation} details the component ablation. Removing explicit Premise Isolation (Step 4) causes the adversarial macro average to drop to 65.1\% (numerically close to Vanilla CoT in Table~\ref{tab:adversarial}, with no statistical difference test reported), suggesting that isolating the specific contradictory sentence is important for the observed diagnostic signal; removing the Divergence Check (Step 3) gives a similar 66.0\% (Table~\ref{tab:ablation}), indicating Step 4 isolation contributes slightly more than Step 3. From the reported macro average and the two displayed cells, the mean of the omitted Temporal Shift and Distractor Evidence cells is about 56.6\% without Step 4 and 56.8\% without Step 3, below Standard RAG on those families (68.8\%/68.8\%); the decomposition scaffold is therefore net-harmful there unless premise isolation and divergence checking are present.

To rule out that the observed CDD signal derives merely from increased generation length, we evaluate a length-matched sham variant: a 5-step prompt structurally similar to CDD but with semantically vacuous steps (e.g., ``restate the question'', ``list entities''). The sham variant obtains a 40.1\% adversarial macro average (Table~\ref{tab:ablation}), below full CDD (78.1\%). This supports the narrower claim that the type of decomposition matters; adding generic multi-step structure is not sufficient and may worsen context-compliance behavior, although parse-failure analysis is needed before attributing the full sham drop to reasoning behavior rather than output-format fragility.

\subsection{Diagnostics \& Compute Tradeoff (CDD-\texorpdfstring{$\alpha$}{alpha})}
\begin{figure*}[t]
\centering
\includegraphics[width=\textwidth]{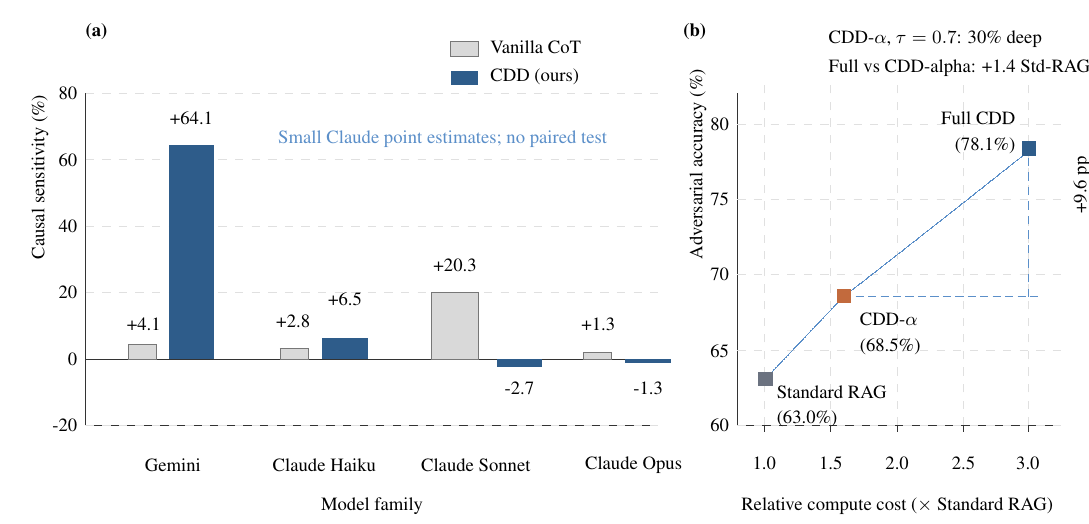}
\caption{Diagnostics and compute trade-off. (a) Mistake-injection causal sensitivity across model families (N=2,250 per cell for all four model cells). On Gemini-2.5-Flash, CDD's final answer is highly sensitive to trace corruption (64.1\%); on all three Claude variants the CDD sensitivities lie in the [-3\%, +7\%] range at N=2,250, with small point-estimate sensitivities and no paired test performed even though adversarial accuracy improves under CDD (Table~\ref{tab:cross_model}). Claude Sonnet and Opus CDD sensitivities are negative point estimates (-2.7\%, -1.3\%), meaning accuracy increased slightly under mistake injection. (b) Compute--accuracy trade-off on the Gemini-2.5-Flash adversarial records. Relative compute is measured in Standard-RAG-equivalent generation calls: Standard RAG = 1, Full CDD $\approx$ 3, and CDD-$\alpha$ at 30\% routing costs $0.7\times1 + 0.3\times3 = 1.6$. CDD-$\alpha$ reaches 68.5\%, about 1.0 pp above random 30\% routing by expectation.}
\label{fig:causal-sensitivity}
\end{figure*}

To analyze the compute tradeoff, we evaluated CDD-$\alpha$ at $\tau=0.7$. The threshold was selected by inspecting prompted contradiction-score histograms on a 100-sample subset drawn from the Epi-Scale adversarial evaluation pool and choosing a value that separated the visible high-conflict cluster from the low-conflict tail; this is threshold-selection on held-out examples from the benchmark, not a separately collected development set. We did not perform a full $\tau$ sweep, which we note as a limitation in the discussion of CDD-$\alpha$. At $\tau=0.7$, CDD-$\alpha$ routed 30.0\% of samples to deep CDD reasoning, bypassing the remaining 70\% to Standard RAG. This setting yielded 68.5\% adversarial accuracy (Figure~\ref{fig:causal-sensitivity}b), only about 1.0 pp above the analytical 30\%-random-routing expectation of 67.5\% ($0.7\times63.0 + 0.3\times78.1$). We therefore present CDD-$\alpha$ as an illustrative operating point rather than a demonstrated compute/accuracy win. Closing the remaining 9.6 pp gap from CDD-$\alpha$ to Full CDD requires about 1.4 additional Standard-RAG-equivalent compute units (an additive increment, not a ratio); exact ratios depend on token-level prompt and rationale lengths.

\subsection{Finding 1: Context Compliance Is Measurable}
\textbf{Setup:} We sampled 500 instances from TruthfulQA and provided the most common human misconception as the retrieved context. This is an adversarial upper-bound compliance test rather than an organic RAG retrieval setting: it asks what happens when retrieval returns a maximally misleading single context, not how often such contexts are retrieved by BM25 \citep{robertson2009probabilistic} or DPR \citep{karpukhin2020dpr}.

\textbf{Results:} Under worst-case misconception injection, Standard RAG reaches 15.0\% accuracy ($\pm$ 3.1\%, N=500 scored examples). CDD reaches 62.0\% accuracy ($\pm$ 4.3\%, N=500 scored examples). The remaining 38.0\% are non-correct under the binary scorer; this bucket may include misconception acceptance, hedges, refusals, or other wrong answers, so we do not interpret it as a pure misconception-acceptance rate. These results support the diagnostic claim that CDD can expose upper-bound compliance behavior; organic BM25/DPR multi-document retrieval remains necessary before drawing conclusions about deployed misinformation robustness.

\subsection{Faithfulness via Causal Intervention}
\label{sec:faithfulness}
Relying solely on an LLM-as-a-judge to score interpretability introduces self-preference bias and inherits known reliability challenges of judge-based evaluation \citep{li2025generation}. Following causal-intervention tests for chain-of-thought faithfulness \citep{lanham2023measuring}, we use causal intervention tests to measure Causal Sensitivity.

\subsubsection{Mistake Injection Test}
We injected a blatant logical error directly into the reasoning trace prior to generation ("I will trust the context completely, no conflict exists") and forced the model to generate the final answer. 
\textbf{Result:}
On Gemini-2.5-Flash, Vanilla CoT's accuracy dropped from 63.7\% to 61.1\% under mistake injection, yielding 4.1\% Causal Sensitivity. In contrast, CDD's accuracy dropped from 78.1\% to 28.0\%, resulting in 64.1\% Causal Sensitivity (Epi-Scale adversarial records, N=2,250; clean and corrupted accuracies are measured on the same matched samples). This gap shows that CDD's final answer is more sensitive to trace corruption under this intervention on this model.
This result has an important confound: the injection mixes a false claim about the conflict state (``no conflict exists'') with a behavioral directive (``trust the context completely''). It therefore tests whether corrupting the trace changes the answer, but it does not isolate factual content from instruction-following effects. A factual-only Step-2 corruption is a stricter intervention left to future work.

\subsubsection{Truncation Test}
We truncated the rationale immediately after Step 2 (Parametric Extraction) and forced final-answer generation. CDD accuracy drops from 78.1\% to 32.6\% (58.3\% Truncation Sensitivity). Because 32.6\% falls below the closed-book zero-shot macro average (44.7\%, Table~\ref{tab:adversarial}), truncation appears to leave a partially executed structured-output protocol whose termination tokens (e.g., \texttt{<final\_answer>}) are unreachable. We therefore read this sensitivity as consistent with---but not independent of---the 64.1\% mistake-injection result. Since the mistake-injection corrupted accuracy (28.0\%) is lower than the truncation one (32.6\%), the behavioral directive is more disruptive than removing Steps 3--5, so 64.1\% likely upper-bounds the trace's causal contribution. The protocol-preserving mistake-injection result remains our primary causal-faithfulness measurement; stricter step-specific tests are future work.

\subsubsection{Cross-Model Faithfulness}
To explore cross-architecture consistency, we applied the Mistake Injection test (N=2,250, the full Epi-Scale adversarial records) to the Claude family.

Figure~\ref{fig:causal-sensitivity}a shows a dissociation between mistake-injection causal sensitivity and adversarial accuracy transfer. Sensitivity is large on Gemini-2.5-Flash (64.1\%), while all three Claude CDD sensitivities fall in the [-3\%, +7\%] range at N=2,250 as point estimates without paired significance tests; nevertheless, adversarial accuracy gains (Table~\ref{tab:cross_model}) transfer positively across the Claude family. Thus, on the basis of these point estimates alone, the Claude-family accuracy gains cannot be attributed to the resolution trace causally driving the answer. This is consistent with, but does not prove, absence of a causal footprint; full paired testing, power, or equivalence analysis is left to future work. Plausible alternatives include increased output length, format-induced calibration, or alignment-training-specific responses to multi-step prompts, which require open-weight replication to disentangle.
The elevated Vanilla-CoT mistake-injection sensitivity on Claude Sonnet (20.3\%) is an isolated high value among the Claude runs; we report it for completeness and do not over-interpret it.

\section{Findings: What CDD Makes Visible}

\subsection{Perturbation-Level Pattern}
The perturbation-level pattern suggests three robustness modes. Entity Swap and Logical Contradiction benefit most because each conflict is localized as a discrete factual claim. Temporal Shift is directionally higher under CDD relative to Standard RAG (71.3\% vs 68.8\%, overlapping CIs), so we make no robustness claim for this perturbation. Distractor Evidence is likewise directionally higher (69.9\% vs 68.8\%, overlapping CIs), so we do not treat the gain as evidence that the trace helps under noisy or partially irrelevant retrieval.

\subsection{Finding 2: Accuracy Transfer Is Not Mechanism Transfer}
As shown in \S5.5.3, accuracy transfer (Table~\ref{tab:cross_model}) and trace-perturbation sensitivity (Figure~\ref{fig:causal-sensitivity}) dissociate: the adversarial-accuracy benefit of explicit decomposition transfers across the Claude family, whereas mistake-injection sensitivity shows no measurable coupling at N=2,250 (CDD sensitivities within [-3\%, +7\%]) outside Gemini-2.5-Flash. We therefore phrase the finding conservatively---what transfers is the accuracy benefit, not a demonstrated causal coupling between the resolution trace and the final answer. This makes CDD a cross-family accuracy intervention in our controlled setting while leaving its mechanism family-specific; the absence of measurable coupling is consistent with, but does not prove, the absence of a causal footprint.

\subsection{Finding 3: Explicit Decomposition Improves Controlled-Conflict Robustness}
As reported in Table~\ref{tab:adversarial}, the clearest controlled-conflict robustness gains occur on Entity Swap and Logical Contradiction, the two localized factual-conflict settings where explicit premise isolation is most directly applicable. This is where CDD separates most clearly from the strongest baseline, Conflict-aware RAG; the table provides the exact values and intervals.

\subsection{Future Work}
Next evaluations are organic BM25/DPR retrieval for TruthfulQA; multi-document retrieval with one false passage among $k$ true passages; factual-only, rationale-swap, and step-specific interventions; and open-weight replication with logit-level prior/posterior divergence. Potential future directions outside the controlled conflict setting include semi-structured graph/text retrieval and adaptive fusion/reranking settings \citep{tao2026grasp}. On the efficiency side, long-context compression and reasoning-pruning methods are methodologically relevant to reducing the overhead of decomposition-style diagnostics \citep{gao2026dspc,jiang2026drp}.

\section{Conclusion}
We used CDD as an inference-time probe for how standard RAG handles epistemic conflict. Final-answer accuracy does not fully capture conflict handling: Standard RAG can enter a measurable context-compliance regime under worst-case misconception injection, explicit decomposition can improve controlled-conflict robustness, and accuracy gains can transfer across model families even when trace-perturbation sensitivity between the stated resolution trace and final answer does not.

These findings position context compliance as a structural RAG behavior requiring systematic diagnostics. A correct answer can hide whether the model rejected a false retrieval, ignored retrieval, or produced a post-hoc rationale; CDD exposes that arbitration by separating contextual belief, parametric belief, premise isolation, and resolution, then testing whether the final answer is sensitive to trace corruption. This trace-perturbation sensitivity is large on Gemini-2.5-Flash; on the Claude family the same trace improves adversarial accuracy without a large point-estimate causal footprint (Gemini 64.1\% vs Claude within [-3\%, +7\%]), with paired significance not established, so the diagnostic should be read as model-specific rather than universal. We will release the Epi-Scale split (2,250 clean + 2,250 adversarial records), prompt templates, evaluation settings, and scoring specifications upon publication; the internal experimental codebase is not released.

\clearpage
\section*{Limitations}
\begin{itemize}[leftmargin=*,labelsep=0.3em,itemsep=0pt,topsep=0pt,parsep=0pt]
    \item \textbf{Cross-Family Generalization}: The clearest causal-sensitivity signal is on Gemini-2.5-Flash (64.1\% mistake-injection sensitivity), with effect magnitude varying across model families. Characterizing the Claude-family behavior---separating architecture, alignment-training, and prompt-sensitivity contributions---is a natural open-weight extension.
    \item \textbf{Causal-Coupling Uncertainty}: On the Claude family, mistake-injection sensitivities lie within [-3\%, +7\%] at N=2,250, versus 64.1\% on Gemini-2.5-Flash. We report these conservatively as point estimates and read them as no measurable coupling in these runs; paired significance, confidence intervals, or an equivalence/power analysis would be needed to move from this reading to a positive claim about the presence or absence of a causal footprint.
    \item \textbf{TruthfulQA Closed-Book Control}: The 15.0\% figure characterizes context-injected behavior under our binary scorer; adding a no-context control on TruthfulQA-500 would further quantify the prior-versus-context gap and is a direct extension.
    \item \textbf{Baseline Prompt Asymmetry}: Because the Standard RAG prompt implicitly encourages context-reliance under adversarial context while CDD instructs explicit conflict checking, we include a conflict-aware prompt baseline to separate decomposition from instruction effects. CDD retains a margin on the localized factual conflicts (Entity Swap, Logical Contradiction) and the margin closes on Temporal Shift and Distractor Evidence; we accordingly read part of the overall gain as attributable to the conflict-checking instruction and part to decomposition.
    \item \textbf{CDD-$\alpha$ Threshold Tuning}: We select the CDD-$\alpha$ threshold $\tau=0.7$ from a 100-sample histogram inspection on the adversarial evaluation pool rather than a separate development set or full sweep; a $\tau \in \{0.3, 0.5, 0.7, 0.9\}$ ablation with a separately fixed gate would be needed before treating the Pareto point as deployment-relevant. We present CDD-$\alpha$ as an illustrative compute/accuracy operating point rather than a tuned deployment setting.
    \item \textbf{TruthfulQA Realism:} Using the explicit top misconception is a deliberate worst-case probe; organic BM25 retrieval would measure average-case degradation and is the natural follow-up.
    \item \textbf{Prior Answer Not Context-Isolated}: In the CDD protocol, Step 2 elicits the parametric answer after the context is visible, so it is a prior reported under the CDD prompt rather than a context-free parametric answer. We therefore treat the belief-revision framing as conceptual; a separate no-context elicitation would tighten the prior/posterior separation and is a direct extension.
    \item \textbf{Conceptual-Only Belief-Revision Framing}: Our belief-revision framing is conceptual rather than a token-level divergence (e.g., JSD); a white-box open-weight instantiation would enable direct quantitative validation of the compliance/resolution distinction.
    \item \textbf{Stricter Faithfulness Tests}: Because mistake injection combines factual content with a behavioral directive, a sharper causal picture would come from factual-only Step-2 corruption, rationale swaps, step-specific corruption, and answer-hidden interventions with contexts that do not reveal the answer; we outline these as the immediate next interventions.
    \item \textbf{Closed-API Drift}: Closed-API behavior may drift under provider updates, and evaluation configuration choices can change benchmark verdicts \citep{li2026safetyreproconfigurationconditionalrankinstability}; version-pinned open-weight replication would remove one major source of variance.
    \item \textbf{Single-Run Closed-API Evaluation}: The reported closed-API results are single-run evaluations at temperature 0.0. Temperature 0.0 does not guarantee deterministic behavior for hosted APIs, so small directional gaps, especially in the Claude-family replication, should be interpreted with caution until repeated runs or open-weight replications are available.
\end{itemize}

\section*{Reproducibility Statement}
Due to API-provider terms and maintenance constraints, we do not release the full internal experimental codebase, which contains non-generalized infrastructure and API-specific wrappers. We will release the Epi-Scale split (2,250 clean + 2,250 adversarial records), prompt templates, evaluation settings, and scoring specifications upon publication; the internal experimental codebase is not released. The release split contains 2,250 clean-context records and 2,250 adversarial-context records, balanced across HotpotQA, Natural Questions, and FEVER and across the four perturbation types.
The API identifiers used in the reported experiments are:
\begin{center}
\footnotesize
\begin{tabular}{@{}ll@{}}
Gemini & \texttt{gemini-2.5-flash} \\
Claude Haiku & \texttt{claude-haiku-4-5-20251001} \\
Claude Sonnet & \texttt{claude-sonnet-4-6} \\
Claude Opus & \texttt{claude-opus-4-6}
\end{tabular}
\end{center}
The Gemini endpoint has an announced discontinuation date of 2026-06-17, so exact replication against the same endpoint may no longer be possible. The Haiku ID is a dated snapshot. From the Claude 4.6 generation onward, a dateless ID such as claude-sonnet-4-6 is a version-pinned snapshot rather than an evergreen alias, so the model weights for a given ID are fixed; however, the surrounding serving infrastructure (e.g., request routing, moderation classifiers, and sampling) may change over time, which we treat as a reproducibility limitation. Experiments were conducted between January and March 2026. Closed-API model behavior may drift on non-pinned dependencies (e.g., moderation layers), which we treat as a reproducibility limitation. All runs use temperature 0.0 and greedy decoding constraints. CDD-$\alpha$ uses a prompted Gemini-2.5-Flash contradiction gate rather than a separate NLI checkpoint; the gate returns a parsed scalar score from a prompt asking whether the context contradicts the hypothesized answer. Dataset construction draws evenly from HotpotQA, Natural Questions, and FEVER; perturbations are generated along four mutually exclusive axes and audited manually on 50 examples per subgroup (92\% valid conflict generation rate). Metrics use normalized string matching with lowercase, punctuation stripping, and alias mapping; FEVER labels are mapped to Boolean support/refute classes in the appendix. Confidence interval computation follows the formulas detailed in \S4.3. Compute resources: All experiments use closed-API inference at standard rate limits; we did not perform local fine-tuning or training.

\bibliography{paper}

\clearpage
\appendix
\begin{figure*}[t]
\centering
\includegraphics[width=0.95\textwidth]{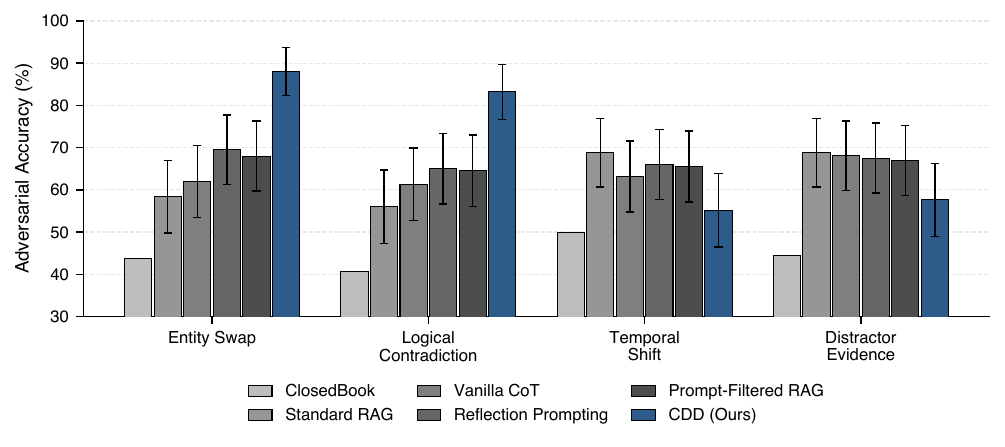}
\caption{Adversarial accuracy on the Epi-Scale adversarial records (Gemini-2.5-Flash, N=2,250; $\sim$562--563 examples per perturbation; normal-approximation 95\% binomial CIs). CDD scores 88.0\%, 83.2\%, 71.3\%, and 69.9\% across the four perturbation types. CDD is highest numerically in each column, with the statistically clearest separation on Entity Swap and Logical Contradiction; the Temporal Shift and Distractor Evidence gains are directional only (overlapping CIs).}
\label{fig:adversarial_accuracy}
\end{figure*}

\section{Appendix: Reproducibility Details}
All evaluations use temperature $0.0$ and greedy decoding constraints. String normalization mapping for FEVER: [supports, true, yes] $\rightarrow$ True; [refutes, false, no, contradicts] $\rightarrow$ False.
\medskip\noindent
\textbf{Standard RAG Template:}\\
\begin{quote}\footnotesize\ttfamily\raggedright
Context: \{context\}\\
Question: \{question\}\\
Based on the context, answer concisely.
\end{quote}
\medskip\noindent
\textbf{ClosedBook Template:}\\
\begin{quote}\footnotesize\ttfamily\raggedright
Question: \{question\}\\
Answer concisely.
\end{quote}
\medskip\noindent
\textbf{Vanilla CoT Template:}\\
\begin{quote}\footnotesize\ttfamily\raggedright
Context: \{context\}\\
Question: \{question\}\\
Think step-by-step, then answer the question. Wrap final answer in \texttt{<final\_answer>}.
\end{quote}
\medskip\noindent
\textbf{Reflection Prompting Template:}\\
\begin{quote}\footnotesize\ttfamily\raggedright
Given the context, answer the question. You must include reflection tokens.\\
Context: \{context\}\\
Question: \{question\}\\
Format your response as:\\
\texttt{[Relevant: Yes/No]}\\
\texttt{[Supported: Fully/Partially/No]}\\
\texttt{[Contradicts\_Prior: Yes/No]}\\
Final Answer: <your concise answer>
\end{quote}
\medskip\noindent
\textbf{Prompt-Filtered RAG Template:}\\
\begin{quote}\footnotesize\ttfamily\raggedright
Context: \{context\}\\
Paraphrase this context. Critically, REMOVE any factual claims that severely violate established world knowledge or history. If a claim seems fake, omit it.\\
Filtered Context: \{filtered\_context\}\\
Question: \{question\}\\
Answer concisely based on the filtered context and your world knowledge. Final Answer:
\end{quote}
\medskip\noindent
\textbf{Conflict-Aware RAG Template:}\\
\begin{quote}\footnotesize\ttfamily\raggedright
Context: \{context\}\\
Question: \{question\}\\
Carefully answer the question using both the retrieved context and your own knowledge.\\
If the retrieved context is consistent with well-established facts, use it as the primary evidence.\\
If the retrieved context appears to conflict with well-established scientific or historical knowledge, explicitly prioritize the more reliable information instead of blindly following the retrieved context.\\
Do not generate intermediate reasoning.\\
Answer concisely.\\
Wrap your final answer in \texttt{<final\_answer>} tags.
\end{quote}
\medskip\noindent
\textbf{Length-Matched Sham CoT Template:}\\
\begin{quote}\footnotesize\ttfamily\raggedright
Context: \{context\}\\
Question: \{question\}\\
Follow these five steps before answering.\\
Step 1: Restate the question in your own words.\\
Step 2: Restate the retrieved context in your own words.\\
Step 3: List the named entities or key terms mentioned in the question and context.\\
Step 4: Identify the broad question type (e.g., person, place, date, fact verification, or explanation).\\
Step 5: Answer the question concisely based on the context.\\
Wrap your final concise answer in \texttt{<final\_answer>} tags.
\end{quote}
\medskip\noindent
\textbf{CDD Template:}\\
\begin{quote}\footnotesize\ttfamily\raggedright
You are resolving cognitive conflicts in retrieval-augmented generation.\\
Context: \{context\}\\
Question: \{question\}\\
Follow these steps:\\
Step 1: Extract the answer implied by the context.\\
Step 2: State your parametric/internal knowledge answer to the question.\\
Step 3: Compare the two answers. Do they conflict?\\
Step 4: If they conflict, isolate the premises causing the conflict. Evaluate if they violate established history/science.\\
Step 5: Resolve the conflict and provide the most factually reliable final answer.\\
Wrap your final concise answer in \texttt{<final\_answer>} tags.
\end{quote}
\medskip\noindent
\textbf{CDD-$\alpha$ Prompted Contradiction Gate:}\\
\begin{quote}\footnotesize\ttfamily\raggedright
Premise: \{context\}\\
Hypothesis: The answer to '\{question\}' is \{parametric\_answer\}.\\
Does the Premise explicitly contradict the Hypothesis? Give a probability from 0.0 to 1.0.
\end{quote}

\section{Appendix: Epi-Scale Details}
\label{sec:appendix_data}
Epi-Scale comprises 2,250 base examples drawn evenly from HotpotQA \citep{yang2018hotpotqa}, Natural Questions \citep{kwiatkowski2019natural}, and FEVER \citep{thorne2018fever}. Each base example is represented as one clean-context record and one adversarial-context record, yielding 4,500 evaluation records. The adversarial records are divided uniformly among Entity Swap, Temporal Shift, Logical Contradiction, and Distractor Evidence. The perturbations were generated using \texttt{gemini-2.5-flash} with prompt templates constraining semantic alterations to exactly one variable axis. To ensure the LLM-based perturbation engine successfully generated genuine conflicts without destroying syntax, we manually audited a random sample of 50 adversarial records per subgroup, finding a 92\% valid conflict generation rate. The audit was conducted as a manual spot check over 200 records total; inter-annotator agreement was not measured, and per-record validity labels were not retained for a filtered-subset sensitivity check.

\end{document}